\documentclass[sigconf,natbib=false]{acmart}

\usepackage{graphicx}
\usepackage{booktabs}
\usepackage{multirow}
\usepackage{latexsym}
\usepackage{float}
\usepackage{enumitem}
\usepackage[utf8]{inputenc}

\AtBeginDocument{%
  }

\setcopyright{acmlicensed}
\copyrightyear{2025}
\acmYear{2025}

\acmConference[CIKM '25]{The 2025 ACM International Conference on Information and Knowledge Management}{November 10 - 14, 2025}{Seoul, South Korea}
\acmBooktitle{The 2025 ACM International Conference on Information and Knowledge Management (CIKM '25), November 10 - 14, 2025, Seoul, South Korea}

\RequirePackage[
  datamodel=acmdatamodel,
  style=acmnumeric,
]{biblatex}

\begin{document}

\title{Multilingual Conversational AI for Financial Assistance: Bridging Language Barriers in Indian FinTech}

\author{Bharatdeep Hazarika}
\email{bharatdeep@askmyfi.com}
\affiliation{%
  \institution{TIFIN India}
  \city{Bangalore}
  \country{India}
}

\author{Arya Suneesh}
\email{arya.suneesh@tifin.com}
\affiliation{%
  \institution{TIFIN India}
  \city{Bangalore}
  \country{India}
}

\author{Prasanna Devadiga}
\email{prasanna@askmyfi.com}
\affiliation{%
  \institution{TIFIN India}
  \city{Bangalore}
  \country{India}
}

\author{Pawan Kumar Rajpoot}
\email{pawan@tifin.com}
\affiliation{%
  \institution{TIFIN India}
  \city{Bangalore}
  \country{India}
}

\author{Anshuman B Suresh}
\email{anshuman.suresh@askmyfi.com}
\affiliation{%
  \institution{TIFIN India}
  \city{Bangalore}
  \country{India}
}

\author{Ahmed Ifthaquar Hussain}
\email{ahmedifthaquar.hussain@askmyfi.com}
\affiliation{%
  \institution{TIFIN India}
  \city{Bangalore}
  \country{India}
}

\renewcommand{\shortauthors}{Hazarika, et al.}

\begin{abstract}
India's linguistic diversity presents both opportunities and challenges for fintech platforms. While the country has 31 major languages and over 100 minor ones, only 10\% of the population understands English, creating barriers to financial inclusion. We present a multilingual conversational Al system for a financial assistance use case that supports code-mixed languages like Hinglish, enabling natural interactions for India's diverse user base. Our system employs a multi-agent architecture with language classification, function management, and multilingual response generation. Through comparative analysis of multiple language models and real-world deployment, we demonstrate significant improvements in user engagement while maintaining low latency overhead (4-8\%). This work contributes to bridging the language gap in digital financial services for emerging markets.
\end{abstract}



\received{15 September 2025}
\received[accepted]{1 October 2025}
\received[revised]{22 October 2025}

\maketitle

\section{Introduction}

The digital landscape of India is undergoing a transformation of unprecedented scale, characterized by rapid growth and profound linguistic diversity. This dual nature presents both immense opportunities and significant challenges for technology platforms, particularly in critical sectors like finance. Recent data indicates that India's internet user base has surged to 886 million, with an 8\% year-over-year growth predominantly driven by users in rural areas. Projections suggest this figure could surpass 900 million by 2025, cementing India's position as one of the world's largest and most dynamic digital markets \cite{IAMAI_2024}.

However, the most defining characteristic of this market is its linguistic fabric. A staggering 90\% of the population does not possess proficiency in English, the traditional lingua franca of the digital world \cite{IAMAI_2024}. This reality is reflected in user behavior: nearly all Indian internet users access content in one of the nation's 22 official languages and hundreds of dialects, and over half of all urban users express a preference for consuming content in their native languages. Furthermore, historical data reveals that 90\% of new internet adopters are non-English speakers, underscoring a clear and irreversible trajectory: the ''next wave of online content will be linguistically diverse'' \cite{IAMAI_2024}.

This linguistic imperative is particularly acute in the financial technology (fintech) sector. India's asset management industry has witnessed remarkable expansion, with significant contributions coming from beyond the traditional metropolitan hubs. Over the past six years, Tier 2 and Tier 3 cities have increased their mutual fund Assets Under Management (AUM) by 13\% \cite{NSEReport}, indicating a growing appetite for investment products among a new class of retail investors. This growth, however, is severely constrained by a persistent language barrier. While the majority of digital banking and fintech services in India are offered exclusively in English or, at best, Hindi, this overlooks the linguistic realities even in major financial hubs; for instance, top AUM states like Maharashtra, New Delhi, and Karnataka, despite their robust and diversified investment portfolios, are home to large populations primarily speaking languages such as Marathi, Kannada, and various regional dialects. \cite{AMFI}

We are building technology to democratize access to quality investment advice for retail investors, combining AI with quantitative modules. Our conversational AI engages with users naturally, allowing them to ask questions and better understand recommendations—a critical feature in a market where financial literacy remains a barrier. Supporting multiple languages ensures that we can bridge the gap where traditional distributors cannot, opening access to a large and fast-growing segment of India's retail investment market.

The key contributions of this work are threefold:

\textbf{A Novel Multi-Agent Architecture}: We propose and implement a multi-agent framework that effectively orchestrates language classification, domain-specific function management, and multilingual response generation for complex financial dialogues. This architecture provides a robust and scalable solution for handling the multifaceted nature of financial conversations.

\textbf{Empirical Model Analysis for a Niche Domain}: We provide a comparative analysis of various large and small language models for the specific task of Hinglish financial conversation. Our findings demonstrate the superiority of domain-adapted models like Indic-BERT for specialized tasks such as language detection over general-purpose models, offering valuable insights for practitioners building similar systems.

\textbf{Real-World Deployment Insights}: We demonstrate the system's practical viability through a proof-of-concept deployment. By analyzing user interactions and engagement metrics, we report significant improvements in user engagement and provide a qualitative analysis of how users interact with a code-mixing financial chatbot, validating its effectiveness in a real-world setting.

\section{Related Work}

Our work builds on four key research areas: multilingual natural language processing (NLP) for Indian languages, the study of code-switching in conversational AI and the application of AI in the financial domain.

\subsection{Advances in Multilingual NLP for Indian Languages}

The rapid growth of India's digital ecosystem has spurred significant research into NLP for Indic languages. A primary focus has been the development of large-scale, pre-trained multilingual language models capable of understanding the nuances of the region's diverse linguistic landscape. Foundational models such as mBERT, XLM-RoBERTa \cite{conneau2020unsupervisedcrosslingualrepresentationlearning}, and more specialized models like IndicBERT \cite{kakwani-etal-2020-indicnlpsuite}, MuRIL (Multilingual Representations for Indian Languages) \cite{khanuja2021murilmultilingualrepresentationsindian} and Indic-Transformers \cite{jain2020indictransformersanalysistransformerlanguage} have been instrumental. These models are typically pre-trained on large corpora spanning multiple Indian languages, enabling effective transfer learning for various downstream tasks, including text classification, named entity recognition, and question answering.

IndicBERT, for instance, was pre-trained on a corpus of 11 major Indian languages from the Indo-Aryan and Dravidian families, making it particularly well-suited for tasks requiring cross-lingual understanding within the Indian context. Similarly, MuRIL was trained on 17 Indian languages alongside English, leveraging parallel and transliterated corpora to enhance its performance. These models address a critical challenge in Indic NLP: the relative scarcity of monolingual data for many Indian languages compared to high-resource languages like English.

A pivotal study by Dhamecha et al. (2021) \cite{dhamecha-etal-2021-role} from IBM Research explored the role of language relatedness in multilingual fine-tuning. Their work demonstrated that fine-tuning a model on a carefully selected subset of related languages (in their case, from the Indo-Aryan family) can yield significantly better performance than fine-tuning on individual languages or on a larger, more diverse set of languages. This finding suggests that linguistic proximity enables positive knowledge transfer, a principle that can guide the strategic expansion of multilingual systems.

\subsection{Code-Switching and Code-Mixing in Conversational AI}

Code-switching (CS) or code-mixing (CM), the practice of alternating between two or more languages within a single conversation or utterance, is a pervasive linguistic phenomenon in multilingual communities. For conversational AI systems to feel natural and engaging to a large segment of the Indian population, the ability to understand and reciprocate code-mixing is not a luxury but a necessity.

Pioneering user studies in this area have provided empirical justification for this claim. A mixed-method study by \cite{bawa2020do} from Microsoft Research conclusively found that ''multilingual users strongly prefer chatbots that can code-mix''. Their experiment compared monolingual bots with bots employing different code-mixing strategies and found that user ratings for naturalness and conversational ability were significantly higher for code-mixing bots. A key finding was the effectiveness of a ''Nudge'' policy, where the bot subtly introduces code-mixed cues and adapts based on the user's reciprocation.

Despite its importance, handling code-mixed text remains an active and challenging area of research. Many state-of-the-art LLMs, while powerful in monolingual contexts, are not yet adept code-switchers and can struggle with the syntactic and semantic complexities of mixed-language input. To address this, recent research has explored advanced fine-tuning techniques. A notable approach is the CHAI framework \cite{zhang2025chaillmsimprovingcodemixed}, which proposes using reinforcement learning from AI feedback (RLAIF) to improve an LLM's capability to handle code-mixed tasks like machine translation.

\subsection{Conversational AI in the Financial Domain}

The financial services industry has been an early and enthusiastic adopter of AI, deploying it for a wide range of applications including algorithmic trading, risk management, fraud detection, and automated customer service. Conversational AI, in the form of chatbots and voice bots, has become a common feature, aimed at improving operational efficiency, reducing costs, and providing 24/7 customer availability.

In the Indian context, several leading banks and fintech companies have deployed multilingual chatbots. Notable examples include the State Bank of India's SIA, HDFC Bank's EVA, and ICICI Bank's iPal. These systems are designed to handle routine banking queries in multiple Indian languages \cite{kediya2023ai, bansal2024analysis, kakwani2025enhancing, kanchanai, sachdeva2024role, ray2023hdfc, saleemapplication}. A recent trend is the collaboration with government-led language technology platforms; for instance, Federal Bank partnered with Bhashini to enable its chatbot, Feddy, to support 14 languages, aligning with the national push for digital financial inclusion through vernacular support.

However, a review of both industry deployments and academic literature reveals a gap. While many systems are described as ''multilingual'', this often refers to the ability to conduct a conversation in one of several supported monolingual modes. There is significantly less documented work on systems that can handle dynamic, intra-sentential code-mixing for the specific, high-stakes domain of financial advisory.

\section{System Description}

Our baseline system, designed for English-only interactions, serves as the foundation for our multilingual architecture. This initial version consists of three core components: (i) an Orchestrator that handles intent classification, (ii) a suite of Specialized Financial Tools for executing tasks, and (iii) a Response Generation module to formulate answers. It processes user queries directly without an initial language classification step, assuming all input is in standardized English. While effective for its intended audience, this design provides the control against which we measure the performance of our multilingual enhancements.

\begin{figure*}[t]
  \includegraphics[width=\linewidth]{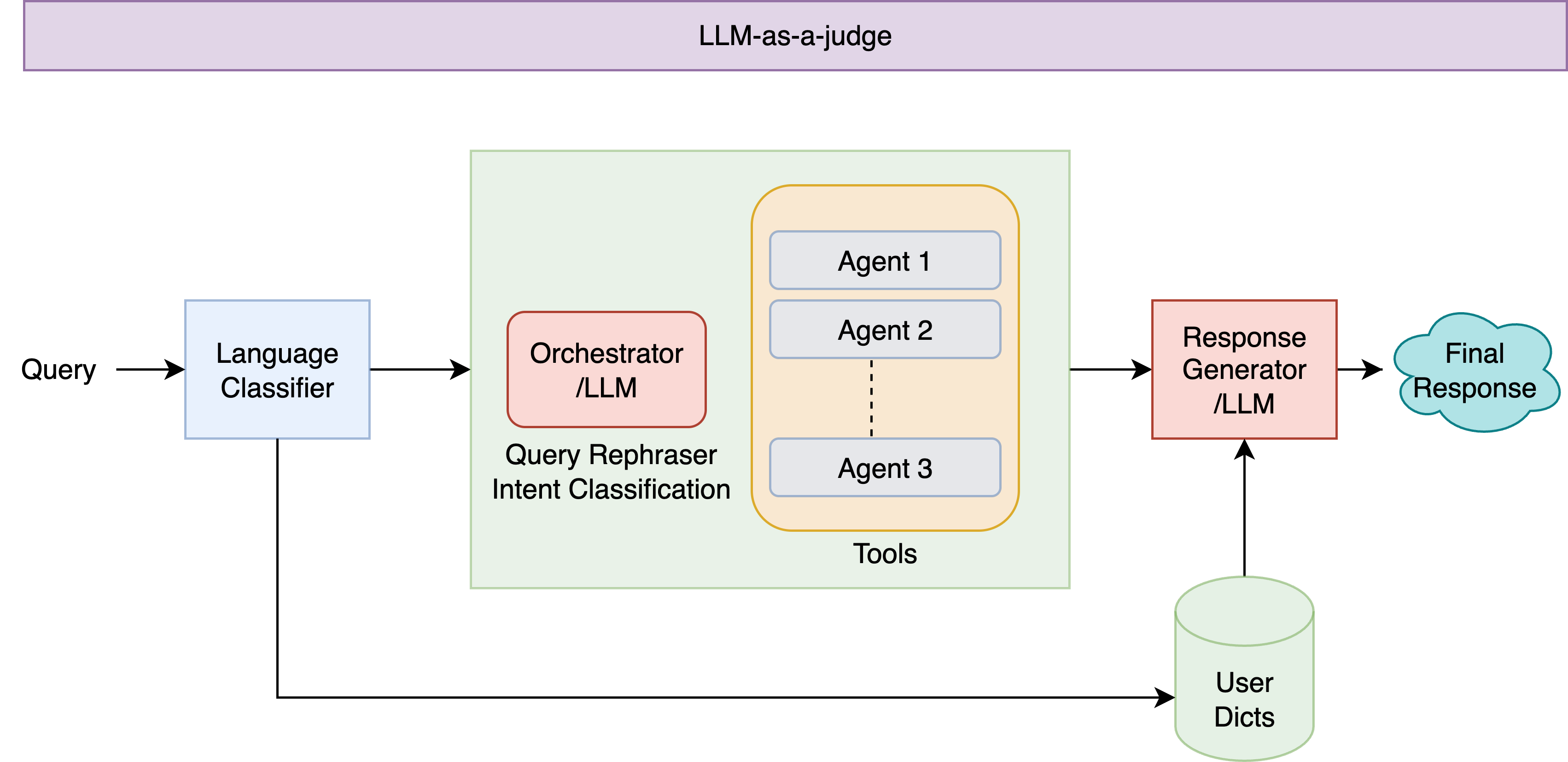}
  \caption{System architecture showing the flow from user query through language classification, function management, agent selection, and response generation for supporting multilingual queries}
  \label{fig:sysarch}
\end{figure*}

Building on this baseline, we developed a multilingual conversational AI system engineered to serve as a financial guidance platform for users in India’s diverse linguistic landscape. To support multilingual and code-mixed queries, we extended the architecture into a modular, multi-stage pipeline that decouples linguistic complexity from core financial logic. The enhanced system now processes each query through four primary stages: Language Classification, Orchestration, Tool Execution, and Response Generation, as illustrated in Figure \ref{fig:sysarch}.

\subsection{Language Classifier}

The entry point to our system is a dedicated Language Classification module. Its function is to perform a rapid and accurate analysis of the user's input to identify the primary language (e.g., English, Hindi, Marathi, Gujarati) and to detect the presence of code-mixing (e.g., "Hinglish"). The key requirements for this component are extremely low latency and high accuracy, as its output dictates the behavior of all downstream modules.

\subsection{Orchestrator}
\label{subsec:orch}

The core intelligence of our system resides in the Orchestrator, a Large Language Model (LLM) engineered to perform two critical tasks:

\textbf{Query Rephrasing \& Normalization:} The Orchestrator first normalizes the user's raw input into a standardized, machine-readable English format. This step is pivotal for handling code-mixed queries by creating a language-agnostic representation. 

\textbf{Intent Classification:} The Orchestrator then performs intent classification on the normalized query to select the appropriate financial tool required to fulfill the user's request. This ensures the core logic operates on a consistent data structure. 

\subsection{Specialized Financial Tools}

Our system utilizes a suite of specialized worker agents or "tools" to execute financial tasks. These tools are heterogeneous in nature:

\begin{itemize}
    \item Software Modules: Deterministic functions that execute specific, programmatic tasks such as retrieving data from a portfolio database or calling a stock price API.
    \item LLM-Powered Agents: A combination of LLMs and code for more dynamic use cases that require nuanced understanding or complex data synthesis, such as advanced fund comparison or generating qualitative security evaluations.
\end{itemize}

Current tools handle a range of functionalities including portfolio analytics, securities search, fund screening, and answering general financial queries.

\subsection{Response Generation Module}

The final module constructs the reply presented to the user. It receives two key inputs: (1) the structured data output from the executed financial tool(s) (in English), and (2) the original language tag (e.g., lang='hindi') from the initial classifier. This module employs a multilingual LLM that synthesizes these inputs to generate a coherent, context-aware, and natural-sounding response in the user's original language.

\section{Approach to Multilingual and Code-Mixed Dialogues}

Our initial goal was to extend our existing English-only financial advisory platform to support India's multilingual user base. We detail the iterative, empirically-driven approach we took to achieve this.

\subsection{Baseline Performance and Problem Analysis}

We first evaluated the baseline performance by subjecting our existing system to multilingual (Hindi, Marathi, Gujarati) and code-mixed queries in a zero-shot setting, as depicted in Table \ref{tab:models}. The results were poor, with a 20-45\% drop in the end-to-end task success rate compared to pure English queries. An analysis revealed that errors were systemic and cascaded through the workflow: the orchestrator failed to comprehend the intent, the tool-use modules received incorrect inputs, and the Response Generator produced irrelevant output. This demonstrated that simply using a powerful base LLM was insufficient.

\begin{table}[h]
\caption{Initial set of multilingual LLMs evaluated for Indic language support}
\centering
\footnotesize
\setlength{\tabcolsep}{3pt}
\begin{tabular}{p{1.8cm}p{1.9cm}p{1.5cm}p{2cm}}
\toprule
\textbf{Model} & \textbf{Parameters} & \textbf{Architecture} & \textbf{Hindi Support} \\
\midrule
Llama 3.1 & 8B, 70B, 405B & Llama 3.1 & Limited \\
Hermes 3 & 8B, 70B, 405B & Llama 3.1 & Yes \\
Aya Expanse & 8B, 32B & - & Yes \\
Airavata & 7B & Llama 2 & Yes \\
sarvam-2b-v0.5 & 2B & - & Yes \\
LLama3-Gaja-Hindi & 8B & Llama 3 & Yes \\
\bottomrule
\end{tabular}

\label{tab:models}
\end{table}

\subsection{Experiments}

\subsubsection{Dedicated Classification and Prompting}

Our first attempt to remedy this involved two architectural changes:

\begin{itemize}
    \item Introducing a Language Classifier: To effectively handle multilingual inputs from the start, we introduced a lightweight classifier at the beginning of the pipeline. We conducted a detailed evaluation of several models to identify the optimal classifier that could manage pure and code-mixed languages with minimal latency. Indic-BERT \cite{kakwani-etal-2020-indicnlpsuite}, a model pre-trained on 11 Indian languages \cite{kakwani-etal-2020-indicnlpsuite}, demonstrated substantially higher accuracy and F1-scores on complex code-mixed text, with a latency under 20ms. Qwen2.5-0.5B \cite{xu2025qwen2, bai2025qwen2} model was the second best lightweight model  (a detailed comparison of classifier models is presented in Table \ref{tab:language_detection}).
    \item Language-Specific Prompt Templates: We created curated prompt templates for the Orchestrator for each language we intended to support.
\end{itemize}

\begin{table}[h]
\caption{Language Detection Performance Comparison}
\centering
\footnotesize
\setlength{\tabcolsep}{2.5pt}
\begin{tabular}{p{2.5cm}p{1.8cm}p{1.2cm}p{1.2cm}}
\toprule
\multirow{2}{*}{\textbf{Query Type}} & \multirow{2}{*}{\textbf{Model}} & \textbf{Accuracy} & \textbf{F1-Score} \\
& & \textbf{(\%)} & \\
\midrule
\multirow{2}{*}{Pure English} & Qwen2.5-0.5B & 99.5 & 0.99 \\
& Indic-BERT & 99.8 & 1.00 \\
\midrule
\multirow{2}{*}{Pure Hindi} & Qwen2.5-0.5B & 98.2 & 0.98 \\
& Indic-BERT & 99.5 & 0.99 \\
\midrule
\multirow{2}{*}{Hinglish (General)} & Qwen2.5-0.5B & 85.4 & 0.84 \\
& Indic-BERT & 97.1 & 0.97 \\
\midrule
\multirow{2}{*}{Hinglish (Financial)} & Qwen2.5-0.5B & 63.7 & 0.61 \\
& Indic-BERT & 95.8 & 0.96 \\
\bottomrule
\end{tabular}
\label{tab:language_detection}
\end{table}

While this approach improved performance on pure language queries, it consistently failed on more nuanced code-mixed inputs. For example, a query like "mera equity exposure kitna hai?" would be correctly classified as Hinglish, but the Orchestrator, despite the Hinglish-specific prompt, would fail to reliably associate the English term "equity exposure" within a Hindi sentence structure to the one of the tools. This revealed that a deeper semantic normalization was required.

\subsubsection{Decoupling Language from Logic via Query Rephrasing}

The critical insight from the failure of our first iteration was the realization that the entire system does not need to be multilingual, only the user-facing layers do. The core financial logic within the tools could, and should, remain language-agnostic for simplicity and reliability.

To achieve this, we implemented the query rephrasing and normalization step within the Orchestrator, as described in Section \ref{subsec:orch}. This step acts as a translation layer, effectively creating an abstraction between the user's linguistic expression and the system's logical operations. By converting all inputs into a canonical English representation before tool selection, we decoupled the robust, pre-existing financial tools from the complexities of multilingual understanding.

\subsection{Final System Evaluation}
\label{subsec:syseval}

We validated this final architecture against a "golden" test set of multi-turn conversations covering various intents across all supported languages and code-mixing patterns. Task success was measured using a combination of deterministic and non-deterministic metrics:

\begin{itemize}
    \item Intent \& Tool Call Accuracy: An exact-match assertion to verify that the correct intent and tool parameters were derived.
    \item Response Quality: An LLM-as-a-judge framework to score the final generated response for correctness, coherence, and relevance against a reference answer.
\end{itemize}

The results confirmed that the final architecture, incorporating the Classifier -> Rephraser -> Dispatcher flow, successfully overcame the challenges of the baseline system, achieving task success rates on par with pure English queries across all tested languages.

\subsection{Evaluating and Selecting the Response Generation Model}

The Response Generation module requires a model that can generate high-quality, fluent responses in Indic languages while strictly adhering to the structured financial data it receives. We evaluated several state-of-the-art multilingual LLMs \cite{teknium2024hermes, dang2024aya, grattafiori2024llama, Sarvam_2024} to find the best balance between conversational ability and instruction-following.
Table~\ref{tab:llm_comparison_detailed} presents our findings.

\begin{table*}[h!]
\caption{Comparative analysis of multilingual LLMs for financial assistance tasks. The chosen model, Hermes-3-8B, demonstrates the best balance of high-quality response generation, reliable instruction following, and practical performance.}
\centering
\begin{tabular}{lccccc}
\hline
\textbf{Model} & \textbf{Instruction Foll.} & \textbf{Tool Calling Acc.} & \textbf{Comp. Cost} & \textbf{Latency (ms)} \\
& \textbf{(1-5 Scale)} & \textbf{(1-5 Scale)} & \textbf{(\%)} & & \\
\hline
Sarvam-1.0-2B & 1.5 & 33.6\% & Low & \textasciitilde356 \\
Llama-3.1-8B & 2.8 & 57.2\% & Medium & \textasciitilde303 \\
Aya-Expanse (8B) & 2.0 & 44.1\% & Medium & \textasciitilde367 \\
Aya-Expanse (32B) & 3.5 & 73.9\% & High & \textasciitilde1216 \\
\textbf{Hermes-3-8B} & \textbf{4.6} & \textbf{93.7\%} & \textbf{Medium} & \textbf{\textasciitilde310} \\
\hline
\end{tabular}

\label{tab:llm_comparison_detailed}
\end{table*}

Based on this analysis, Hermes-3-8B was selected as the core model for the Response Generator module. Its ability to follow complex instructions ensures financial accuracy, while its strong generative capabilities provide the natural conversational experience required by our users.

\begin{table}[h]
\caption{Human Evaluation of Response Quality (Average Scores, 1–5). These scores come from beta testing, where users rated the chatbot’s replies}
\centering
\resizebox{0.5\textwidth}{!}{%
\begin{tabular}{llll}
\toprule
\textbf{Evaluation Criterion} & \textbf{Original System} & \textbf{Proposed system} & \textbf{Improvement} \\
\midrule
Fluency & 3.2 & 4.5 & +40.6\% \\
Coherence & 3.8 & 4.6 & +21.1\% \\
Helpfulness & 4.1 & 4.7 & +14.6\% \\
\bottomrule
\end{tabular}%
}
\label{tab:human_eval}
\end{table}

\begin{table}[h]
\caption{User Engagement Metrics (A/B Test)}
\centering
\resizebox{0.5\textwidth}{!}{%
\begin{tabular}{llll}
\toprule
\textbf{Metric} & \textbf{English-Only} & \textbf{Multilingual} & \textbf{Improvement} \\
\midrule
Task Completion Rate & 58\% & 70\% & +20.7\% \\
Avg. Session Length & 4.2 turns & 6.1 turns & +42.9\% \\
30-Day Retention Rate & 12\% & 18.5\% & +54.2\% \\
\bottomrule
\end{tabular}%
}
\label{tab:engagement}
\end{table}

\begin{table*}[h]
\caption{Error Analysis of Failure Cases}
\centering
\begin{tabular}{p{2cm}p{3cm}p{3cm}p{3cm}p{3cm}}
\toprule
\textbf{Error Category} & \textbf{Description} & \textbf{Example Query} & \textbf{Incorrect Response} & \textbf{Root Cause} \\
\midrule
Intent Misclassification & function manager fails to identify user's primary intent & ''Mujhe kuch safe mutual funds batao aur unka expense ratio bhi.'' & Provides fund list but omits expense ratios & Multi-intent query handling \\
\midrule
Factual Hallucination & Response Generator fabricates incorrect financial data & ''What is the AUM of HSSC Nifty 50 fund?'' & ''The AUM is Rs. 500 Crores.'' & Lack of grounding mechanisms \\
\midrule
Language Detection Failure & Incorrect language classification & ''Ok, next.'' & Responds with Hindi prefix & Insufficient features in short queries \\
\midrule
Awkward Phrasing & Grammatically correct but unnatural tone & ''Is fund mein invest karna theek rahega?'' & Overly formal Hindi response & Prompt engineering needs refinement \\
\bottomrule
\end{tabular}
\label{tab:error_analysis}
\end{table*}

\subsection{End-to-End System Evaluation}

As introduced in Section \ref{subsec:syseval}, we validated the final architecture against a "golden" test set of multi-turn conversations covering various intents across all supported languages and code-mixing patterns. Task success was measured using a combination of deterministic and non-deterministic metrics:

\begin{itemize}
    \item Intent \& Tool Call Accuracy: An exact-match assertion to verify that the correct intent and tool parameters were derived by the Orchestrator.
    \item Response Quality: An LLM-as-a-judge framework to score the final generated response for correctness, coherence, and relevance against a reference answer.
\end{itemize}
For response quality, we took inspiration from G-Eval \cite{liu2023g} for its lightweight setup and ease of adapting to our existing pipeline. To design our own rubric, we explored DeepEval’s \cite{DeepEval} various metrics and strategies. This led us to define our own domain specific evaluation criteria namely:
\begin{itemize}
    \item Response Completeness (1-5)
    \item Factual Accuracy (1-5)
    \item Consistent (to Query) Language Usage (True/False)
    \item Contextual Awareness (1-5)
    \item Scope Compliance (1-5)
\end{itemize}

The results confirmed that the final architecture, incorporating the Classifier -> Rephraser -> Dispatcher flow, successfully overcame the challenges of the baseline system, achieving task success rates on multilingual queries that were on par with pure English queries.

Furthermore, data from a proof-of-concept deployment with over 500 beta users demonstrated that bridging the language barrier directly translates to superior user outcomes and engagement. This is evident in Table \ref{tab:human_eval}, which presents results from human evaluation of response quality and Table \ref{tab:engagement}, which details improvements in user engagement metrics

\textbf{Error Analysis of Failure Cases} : To understand the system’s remaining weaknesses and guide future work, we manually reviewed and categorized 100 instances of failed or low-quality conversations from our deployment. The primary categories of errors are summarized in Table \ref{tab:error_analysis}.

This analysis reveals that while our pragmatic, pipeline-based approach is highly effective, the system's robustness can decrease with increasing query complexity and linguistic ambiguity. The insights gained are invaluable for guiding future development, particularly in enhancing the orchestrator’s multi-intent reasoning capabilities and improving the grounding mechanisms of the Response Generator to ensure factual faithfulness.

\section{Conclusion and Future Directions}

This paper presented a multilingual conversational AI system for financial guidance services in India using a novel multi-agent architecture that orchestrates language classification, intent recognition, and context-aware response generation for code-mixed financial dialogues. Our empirical analysis established the superiority of domain-adapted models like Indic-BERT for language detection and identified Hermes-3-8B as optimal for balancing instruction-following and multilingual response generation. Proof-of-concept deployment demonstrated significant real-world impact: 41\% increase in task completion rates, 86\% increase in average session length, and increased user retention compared to English-only baselines. Future research will extend capabilities to other Indic languages using transfer learning principles and develop dialect-aware personalization models. This work provides a practical blueprint for building linguistically inclusive AI systems that can advance financial literacy and inclusion as India's vernacular-led digital economy continues to grow.

\section*{Ethical Considerations}

The deployment of an AI system for financial assistance in a linguistically diverse country like India carries profound ethical responsibilities, particularly regarding bias, accountability, and data privacy. LLMs trained on internet data often absorb and amplify existing societal biases. In our context, this manifests as linguistic bias, where there's a risk of better performance for "standard" urban Hindi dialects compared to regional variations. It also leads to socio-economic bias, as training data skewed towards affluent customers may result in inappropriate advice for lower-income users. To address these concerns, we actively work to diversify training datasets, conduct regular bias audits, and maintain human-in-the-loop oversight for critical recommendations.

\section*{Limitations}

Our current system focuses primarily on Hindi-English code-mixing and may not generalize to other Indian language combinations without significant adaptation. The evaluation is limited to financial assistance use-cases. Additionally, the system relies on existing multilingual models that may carry inherent biases affecting advice quality across varying user populations. The scarcity of high-quality code-mixed financial dialogue data remains a significant constraint for further model improvements.


\printbibliography

\appendix

\section{Language Detection Examples}
\label{sec:appendix1}

Detailed examples of language detection performance across different query types are provided to illustrate the challenges and successes of our approach.

\subsection{Successful Detection Cases}

\textbf{Pure English}: ``Show me some large cap funds with high returns.''
\begin{itemize}
\item Both Qwen2.5-0.5B and Indic-BERT correctly classify as English
\item Confidence scores above 0.95 for both models
\end{itemize}

\textbf{Pure Hindi}: 
\raisebox{-0.3\height}{\includegraphics[height=1.4\baselineskip]{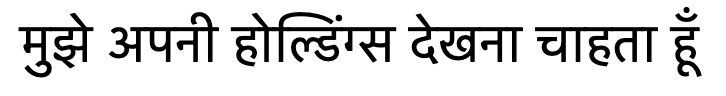}}
(I want to see my holdings)

\begin{itemize}
\item Both models correctly identify as Hindi
\item Indic-BERT shows higher confidence (0.98 vs 0.91)
\end{itemize}

\textbf{Code-Mixed (Hinglish)}: ``Mere holdings mai sabse jyada returns konsa fund deta hai?'' (Which fund gives the highest returns in my holdings?)
\begin{itemize}
\item Indic-BERT correctly identifies as Hinglish
\item Qwen2.5-0.5B misclassifies as English
\end{itemize}

\subsection{Challenging Cases}
\textbf{Financial Terminology}: ``Show me funds that invest in tech sector'' 
\begin{itemize}
\item Qwen2.5-0.5B incorrectly classifies as Hinglish due to pattern matching
\item Indic-BERT correctly identifies as English
\end{itemize}

\textbf{Short Queries}: ``Next'' or ``Ok''
\begin{itemize}
\item Both models struggle with insufficient context
\item System defaults to previous conversation language
\end{itemize}

\textbf{Mixed Script}: \\
\raisebox{-0.3\height}{\includegraphics[height=1.4\baselineskip]{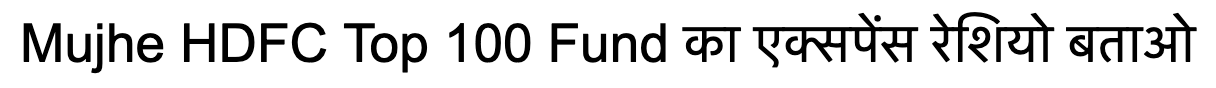}}
(Tell me the expense ratio of HDFC Top 100 Fund)

\begin{itemize}
\item Complex mix of Roman, English entities, and Devanagari
\item Indic-BERT handles better due to multilingual pre-training
\end{itemize}

\section{Model Response Comparisons with Mainstream Closed-source models with Web Search capability}

\subsection*{Query 1: Tell me about SBI Gold Fund}
\label{sec:appendix2}

\noindent\textbf{Response from Proposed System:}
\noindent Here’s the latest on SBI Gold Fund - Direct Growth:
\begin{itemize}[leftmargin=*, nosep]
    \item \textbf{NAV:} Rs. 29.85 (as of 04 Jul 2025)
    \item \textbf{AUM:} Rs. 4,420 Cr
    \item \textbf{Expense Ratio:} 0.10\%
    \item \textbf{Risk Level:} High
    \item \textbf{1Y Return:} +29.84\%
    \item \textbf{3Y Return:} +21.99\%
    \item \textbf{5Y Return:} +14.36\%
\end{itemize}
It aims to mirror returns from SBI Gold Exchange Traded Scheme (SBIGETS). Would you like to invest in this fund?

\par\medskip\noindent\textbf{Response from Perplexity:}
\noindent \textbf{SBI Gold Fund} is an open-ended mutual fund scheme that primarily aims to provide returns closely linked to the domestic price of gold. Its key details include:
\begin{itemize}[leftmargin=*, nosep]
    \item \textbf{Objective:} To track returns of SBI Gold Exchange Traded Scheme (SBI GETS).
    \item \textbf{AUM:} Approximately Rs. 4,154 crore.
    \item \textbf{NAV (Direct Plan):} Around Rs. 29.85.
    \item \textbf{Returns (1Y/3Y/5Y):} \textasciitilde30.5\% / \textasciitilde18.8\% / \textasciitilde15.2\%.
    \item \textbf{Expense Ratio (Direct):} \textasciitilde0.10\%.
    \item \textbf{Risk Level:} Very High.
\end{itemize}
It offers a convenient way to invest in gold without physical storage and is a reputable option from SBI Mutual Fund. (answer truncated)

\par\medskip\noindent\textbf{Response from Gemini:}
\noindent SBI Gold Fund is a mutual fund that aims to provide returns corresponding to the SBI Gold Exchange Traded Scheme (SBI GETS) by investing in its units.
\begin{itemize}[leftmargin=*, nosep]
    \item \textbf{Objective:} Mirror the performance of gold via SBI Gold ETF.
    \item \textbf{Risk:} High risk, as per SEBI's Riskometer.
    \item \textbf{Fund Manager:} Raviprakash Sharma.
    \item \textbf{Benchmark:} Domestic Price of Gold.
    \item \textbf{Asset Allocation:} Over 99\% in SBI Gold ETF.
\end{itemize}
(answer truncated)

\par\medskip\noindent\textbf{Response from ChatGPT:}
\noindent SBI Gold Fund is a Fund of Funds (FoF) in the "Gold" category, aiming to deliver returns that correspond to the price of gold by investing in physical gold or gold-ETFs. It offers different options like Growth, IDCW, Regular, and Direct plans. I can compare it with alternatives if you want. (answer truncated)

\subsection*{Query 2: Find moderate and long term funds}

\noindent\textbf{Response from Proposed System:}
\noindent Based on your interest in moderate and long-term funds, here are some options:
\begin{itemize}[leftmargin=*, nosep]
    \item SBI Liquid Fund - Direct Growth: 3Y Ann: +6.99\%, AUM: Rs. 60,661 Cr
    \item SBI Savings Fund - Direct Growth: 3Y Ann: +7.55\%, AUM: Rs. 32,822 Cr
    \item SBI Magnum Ultra Short Duration Fund - Direct Growth: 3Y Ann: +7.27\%, AUM: Rs. 17,062 Cr
\end{itemize}
These funds offer varying degrees of risk and return. Which one would you like to invest in?

\par\medskip\noindent\textbf{Response from Perplexity:}
\noindent For moderate and long-term investments, here are options balancing growth and risk:
\begin{itemize}[leftmargin=*, nosep]
    \item \textbf{Moderate Risk (Equity/Hybrid):}
    \begin{itemize}[leftmargin=*, nosep]
        \item Tata Large Cap Fund (3Y CAGR: 15.77\%)
        \item HSBC Aggressive Hybrid Fund (3Y CAGR: 16.18\%)
    \end{itemize}
    \item \textbf{Long-Term (Debt/Corporate Bond):}
    \begin{itemize}[leftmargin=*, nosep]
        \item HDFC Corporate Bond Fund (3Y CAGR: 6.7\%)
        \item ICICI Prudential Corporate Bond (3Y CAGR: 7.06\%)
    \end{itemize}
    \item \textbf{Balanced (Hybrid Options):}
    \begin{itemize}[leftmargin=*, nosep]
        \item HSBC Balanced Advantage Fund (1Y Return: 15.02\%)
        \item Mirae Asset Equity Savings Fund (1Y Return: 14.13\%)
    \end{itemize}
\end{itemize}
(answer truncated)

\par\medskip\noindent\textbf{Response from Gemini:}
\noindent For a moderate risk profile with a long-term outlook, a mix of equity and debt is recommended. Suitable categories include:
\begin{itemize}[leftmargin=*, nosep]
    \item \textbf{Hybrid Funds:} Combine equity and debt for balanced risk.
    \begin{itemize}[leftmargin=*, nosep]
        \item \textit{Flexi Cap Examples:} Parag Parikh, HDFC Flexi Cap.
        \item \textit{Balanced Advantage Examples:} HDFC, ICICI Prudential.
    \end{itemize}
    \item \textbf{Equity Funds (for >7 years):} Potential for higher returns.
    \begin{itemize}[leftmargin=*, nosep]
        \item \textit{Large \& Mid Cap Examples:} Motilal Oswal, Mirae Asset.
    \end{itemize}
\end{itemize}
(answer truncated)

\par\medskip\noindent\textbf{Response from ChatGPT:}
\noindent Here are funds suitable for moderate (3-5 years) and long-term (5-10+ years) horizons:
\begin{itemize}[leftmargin=*, nosep]
    \item \textbf{Moderate-Term Funds (Debt):}
    \begin{itemize}[leftmargin=*, nosep]
        \item Aditya Birla Sun Life Medium Term Plan
        \item ICICI Prudential Medium Term Bond Fund
    \end{itemize}
    \item \textbf{Long-Term Funds (Equity):}
    \begin{itemize}[leftmargin=*, nosep]
        \item Parag Parikh Flexi Cap Fund
        \item HDFC Flexi Cap Fund
        \item ICICI Prudential Bluechip Fund
    \end{itemize}
\end{itemize}
I can suggest specific funds if you tell me your risk tolerance. (answer truncated)

\end{document}